\documentclass[letterpaper]{article}

\usepackage{times}
\usepackage{helvet}
\usepackage{courier}
\usepackage{amsmath}
\usepackage{float}
\usepackage[utf8]{inputenc}
\usepackage{algorithm}
\usepackage{algorithmic}
\usepackage{graphicx}
\usepackage{tikz}
 
\begin{document}
%
\title{Bangla AI: A Framework for Machine Translation Utilizing Large Language Models for Ethnic Media}
\author{
MD Ashraful Goni*, Fahad Mostafa**, Kerk F. Kee* \\
\small{\texttt{mgoni@ttu.edu}, \texttt{fahad.mostafa@ttu.edu}, \texttt{kerk.kee@ttu.edu}} \\
\small{*College of Media \& Communication, Texas Tech University, USA.}\\
\small{**Department of Mathematics and Statistics, Texas Tech University, USA.}
}
\maketitle
\begin{abstract}
\begin{quote}
Ethnic media, which caters to diaspora communities in host nations, serves as a vital platform for these communities to both produce content and access information. Rather than utilizing the language of the host nation, ethnic media delivers news in the language of the immigrant community. For instance, in the USA, Bangla ethnic media presents news in Bangla rather than English. This research delves into the prospective integration of large language models (LLM) and multi-lingual machine translations (MMT) within the ethnic media industry. It centers on the transformative potential of using LLM in MMT in various facets of news translation, searching, and categorization. The paper outlines a theoretical framework elucidating the integration of LLM and MMT into the news searching and translation processes for ethnic media. Additionally, it briefly addresses the potential ethical challenges associated with the incorporation of LLM and MMT in news translation procedures.
\end{quote}
\end{abstract}

\section{Introduction}

This research paper proposed an algorithm for the uses of the Large Language model (LLM) and multi-lingual machine translations (MMT) in language translation and news searching for ethnic media journalists in the USA, with a specific focus on the Bangladeshi ethnic media in New York City. Ethnic media is a media for the diaspora community in a host nation where the diaspora community utilizes it to produce content for their community  \cite{riggins1992ethnic}. Ethnic media is also known as “minority media, immigrant media, diasporic media, and community media” \cite{matsaganis2018ethnic}. This paper \cite{riggins1992ethnic}explains the dual role of ethnic media, which serves as an ethnic community's access to mainstream and diaspora values. In this sense, we can easily say that ethnic media is also local media because it focuses on the local life of a specific community in a host nation. For example, a Bengali ethnic media based in New York City serves the Bengali diaspora community with information from the homeland; at the same time, they also provide information related to the local community in New York City, such as government benefits, jobs, and many other events that is relevant to the ethnic community.

Some large language models such as GPT-4, and ChatGPT \cite{achiam2023gpt} are useful to translate language and create content in different languages \cite{zhu2023multilingual}. Bangla AI is a project aimed at transforming the Bengali ethnic media landscape using the power of LLM and MMT. The vision of Bangla AI is to empower US-based Bengali journalists with seamless access to critical news and information in their native language, bridging the gap between mainstream media and ethnic media through innovative news searching and translation solutions.

According to a Pew Research Center report, in 2019, 208,000 legal Bangladeshi immigrants lived in the United States. Among these populations, 93,000 live in New York City. The report suggested that only 55\% of Bangladeshi immigrants speak English fluently\cite{budiman_bangladeshis_2019}. This number is higher for illegal Bangladeshi immigrants. Other immigrant communities also have similar statistics related to fluency in English. For example, according to the Pew Research Center, 57\% of lawful US immigrants are fluent in English, while 34\% of unauthorized immigrants are fluent in English \cite{passel_us_2019}. Findings of this paper \cite{ramasubramanian2017ethnic} suggest that the language barrier is one of the major reasons for the consumption of ethnic media in the United States. 

Ethnic media serves as an ethnic community's access to mainstream media information. This media focuses on the local events and news related to the diasporic community \cite{riggins1992ethnic}. For example, content analysis on Bangladeshi ethnic media based in the US finds that Bengali diasporic media provides more information about the ethnic community in New York City than at home (Bangladesh) \cite{biswas2018assessment}. To provide U.S. local news in the ethnic media, journalists often utilize language translation tools such as Google Translate. However, traditional translation tools like Google Translate often fail to deliver contextually accurate translations, especially for large chunks of text and low-resource languages \cite{zhu2023multilingual, zhang2023prompting}. Emerging large language models (LLM) exhibit the potential to overcome these limitations and, in the future, could provide better translation capabilities and generate content in diverse languages through machine learning and machine translation \cite{kreutzer2018can, zhang2023prompting, zhu2023multilingual}. 

Given these circumstances, the Bangla AI algorithm promises to revolutionize ethnic media journalism practices in the USA by catering specifically to marginalized populations and enhancing their access to information. Representing a novel approach, it offers a comprehensive framework for the entire news production process, encompassing every step from news selection to final production.

\section{Methodology}

The procedure begins by collecting multivariate data from different news sources on themes such as employment, immigration, future goals, housing, healthcare, and politics. This data is displayed as a matrix, with each row representing a news article and each column representing the existence or significance of a particular topic. Following data gathering, a classification algorithm is used to divide news articles into several categories based on their content \cite{cooley1999classification, miao2018chinese}. This entails training a model using multivariate data to predict the news type/class for each article. Once trained, the categorization model can correctly assign news articles to their appropriate categories. Furthermore, to enhance accessibility for Bangla-speaking audiences, the specified news pieces can be translated from English to Bangla utilizing recently developed technology. Here a framework has been shown step by step.

\subsection{Data Collection using NLP}

The approach begins with the collection of multivariate data from $p$ different news sources on a wide range of issues $q$, that are relevant to the Bangladeshi immigrant community in New York City including employment, immigration, plans, housing, healthcare, and politics \cite{biswas2018assessment}. This data collection phase uses Natural Language Processing (NLP) \cite{cambria2014jumping, vicari2021analysis} techniques to extract useful information from news stories from authentic news sources such as the New York Times, New York Post, Documented, etc. We will only collect data that allow us to republish their content under copyright and give permission to publish in different language in different ethnic media \cite{marks2018can, johnstone2004debunking}. This approach will help mainstream English news outlets reach a wider demographic audience and break the language barrier.  Google search engine and other social media platforms are leaders in spreading false news, misinformation, and disinformation \cite{metaxa2017google, shao2018anatomy}. Additionally, implementing these gatekeeping measures will aid in mitigating the dissemination of misinformation and disinformation originating from this dataset.

NLP algorithms analyze the textual content of each article to detect and quantify the presence of various themes in the dataset. Following that, the processed data is organized into a matrix format, with rows representing individual news stories and columns representing distinct thematic groups. This organized dataset serves as the framework for subsequent analysis and classification activities, providing insights into the current trends and patterns across a variety of news sources \cite{entman2005nature}.
\subsection{Data Representation}

Let $X$ be an $N \times M$ matrix representing the multivariate data from $p$ different news media on $q$ different topics. Each row of $X$ corresponds to a news article, and each column represents a topic. Mathematically, we can represent $X$ as:

$$X = x_{ij},$$
where $x_{ij}$ denotes the relevance or occurrence of topic $j$ in news article $i$. Additionally, let $Y$ be a vector of length $N$ representing the news type/class for each news article. Mathematically, $Y$ can be represented as:

$$Y = [y_1, y_2, \ldots, y_N],$$
where $y_i$ represents the class label for news article $i$.

\subsection{Classification}

After capturing the multivariate data, we train a classification algorithm to predict the news type/class based on the features represented in $X$. Let's denote the trained classifier as $f(X)$. The classification process can be represented as:

\begin{equation}
    Y_{\text{pred}} = f(X)
\end{equation}
where $Y_{\text{pred}}$ represents the predicted class labels for the news articles.

\subsection{Translation}

To translate the specific news from English to Bangla, we utilize a Large Language Model (LLM) such as GPT-3.5/4 or BERT. Let's denote the translation function as $T(\text{English})$, which takes English text as input and generates the corresponding Bangla translation. Mathematically:

\begin{equation}
    \text{Bangla} = T(\text{English})
\end{equation}

where $\text{Bangla}$ represents the translated text in Bangla. This mathematical formulation depicts the process of extracting multivariate data from news items, categorizing them, and translating the specific news from English to Bangla utilizing LLM models.

\begin{algorithm}{
\caption{ Bangla News search and Translation}\label{alg:news_analysis}
\begin{algorithmic}[1]
\STATE \textbf{Input:} 
\STATE \quad $N$ - Number of news articles
\STATE \quad $M$ - Number of news topics
\STATE \quad $X$ $\in$ $N \times M$ matrix capturing news features
\STATE \quad $Y$ - Vector of length $N$ representing news classes
\STATE \quad $C$ - Number of news classes

\STATE \textbf{Data Representation:}
\STATE \quad $X$ - Matrix capturing features of news articles
\STATE \quad $Y$ - Vector representing news classes

\STATE \textbf{Classification:}
\STATE \quad Train a classification algorithm (e.g., logistic regression, neural network) using $X$ and $Y$
\STATE \quad Obtain a trained model for predicting news type/class

\STATE \textbf{Translation:}
\STATE \quad Utilize a Large Language Model (LLM) to translate English news text to Bangla
\STATE \textbf{Output:}
\STATE \quad Translated news articles in Bangla
\end{algorithmic}}
\end{algorithm}

\begin{figure*}[ht!]
  \centering
  \includegraphics[width=0.95\textwidth]{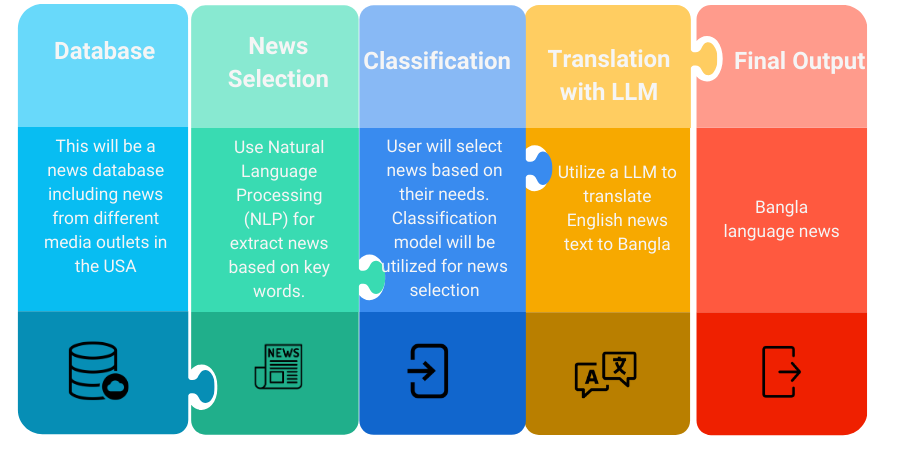}
  \caption{Flowchart of Bangla News Translation for Ethnic Media.}
  \label{fig:flowchart}
\end{figure*}

\section{Discussion and Conclusions}
Several studies have proposed solutions based on LLM and MMT for language translation with context and cultural awareness \cite{lyu2023new,koshkin2024transllama}. Additionally, other research efforts have utilized LLM for detecting fake news and misinformation \cite{leite2023detecting, kareem2023fighting, hu2023bad}. However, to date, no studies have presented a comprehensive understanding of how LLM and MMT can be leveraged to provide better information to marginalized communities while simultaneously combating misinformation and fake news. Our algorithm and flowchart offer a complete solution for translating news from authentic sources using LLM and MMT.

This study is one of the pioneer intellectual approaches to understanding the uses of LLM and MMT in ethnic media or marginalized communities. Crucially, LLM  and MMT are currently in the developmental phase \cite{zhu2023multilingual, zhang2023prompting}. Therefore, this study opens the intellectual door to analyze and comprehend the potential of these technologies. This futuristic study will help us to develop a comprehensive policy to address ethical considerations and potential challenges before the widespread adoption of LLM and MMT in the ethnic media industry. Mainstream media can play a significant role in improving the quality of ethnic media. However, at the same time, it is essential to understand the mutual interplay rather than complete domination from mainstream media. There should be a flow from both media and dependency on both sides. 

To conclude, Bangla AI empowers journalists to ensure comprehensive and accurate news coverage. This breakthrough not only connects mainstream English media with ethnic media but also ensures that crucial information reaches the Bengali-speaking audience, strengthening the role of ethnic media as a trusted source of news. The broader impact of this study is not only limited to the Bangladeshi community rather it can be helpful for all other ethnic communities in the USA. By employing LLM and MMT, other ethnic media houses in the USA (e.g. Hispanic ethnic media, Indian ethnic media, and Chinese ethnic media) can effectively connect with diverse audiences and overcome language barriers. 


\end{document}